\documentclass[conference]{IEEEtran}
\IEEEoverridecommandlockouts

\usepackage{cite}
\usepackage{amsmath,amssymb,amsfonts}
\usepackage{algorithmic}
\usepackage{graphicx}
\usepackage{textcomp}
\usepackage{xcolor}
\usepackage{multirow}
\usepackage{booktabs}
\usepackage{hyperref}
\usepackage{amsmath}
\usepackage{amsfonts}

\newcommand{\E}{\mathbb{E}}
\newcommand{\Enc}[1]{\text{Enc}\left(#1\right)}
\newcommand{\T}[1]{\text{T}\left(#1\right)}

\def\BibTeX{{\rm B\kern-.05em{\sc i\kern-.025em b}\kern-.08em
    T\kern-.1667em\lower.7ex\hbox{E}\kern-.125emX}}
\begin{document}

\title{Flow-SLM:\\ Joint Learning of Linguistic and Acoustic Information for Spoken Language Modeling\\

}

 \author{\IEEEauthorblockN{Ju-Chieh Chou}
 \IEEEauthorblockA{
\textit{Toyota Technological Institute at Chicago}\\
IL, USA \\
jcchou@ttic.edu}
\and
\IEEEauthorblockN{Jiawei Zhou}
\IEEEauthorblockA{
\textit{Stony Brook University}\\
NY, USA \\
jiawei.zhou.1@stonybrook.edu}
\and
\IEEEauthorblockN{Karen Livescu}
\IEEEauthorblockA{
\textit{Toyota Technological Institute at Chicago}\\
IL, USA \\
klivescu@ttic.edu}
}

\maketitle
\begin{abstract}

Textless spoken language models (SLMs) are generative models of speech that do not rely on text supervision.  Most textless SLMs learn to predict the next semantic token, a discrete representation of linguistic content, and rely on a separate vocoder to add acoustic information to the generated speech.  Such models have no access to acoustic context and no built-in control over acoustic details.
In this work, we propose to jointly model linguistic and acoustic information by generating semantic tokens and a continuous real-valued representation of the acoustic frame.  We use a flow-matching objective to predict the continuous vector conditioned on the semantic tokens.
We study the design space of this approach and find that predicting multiple future semantic tokens helps preserve linguistic information. Our approach achieves comparable performance to existing models in terms of linguistic likelihood benchmarks, while providing better acoustic detail in prompted generation.\footnote{Project page: \url{https://jjery2243542.github.io/flowslm.github.io/}}
\end{abstract}

\begin{figure*}
    \centering
    \hspace{.05\linewidth}
    \includegraphics[width=.35
    \linewidth]{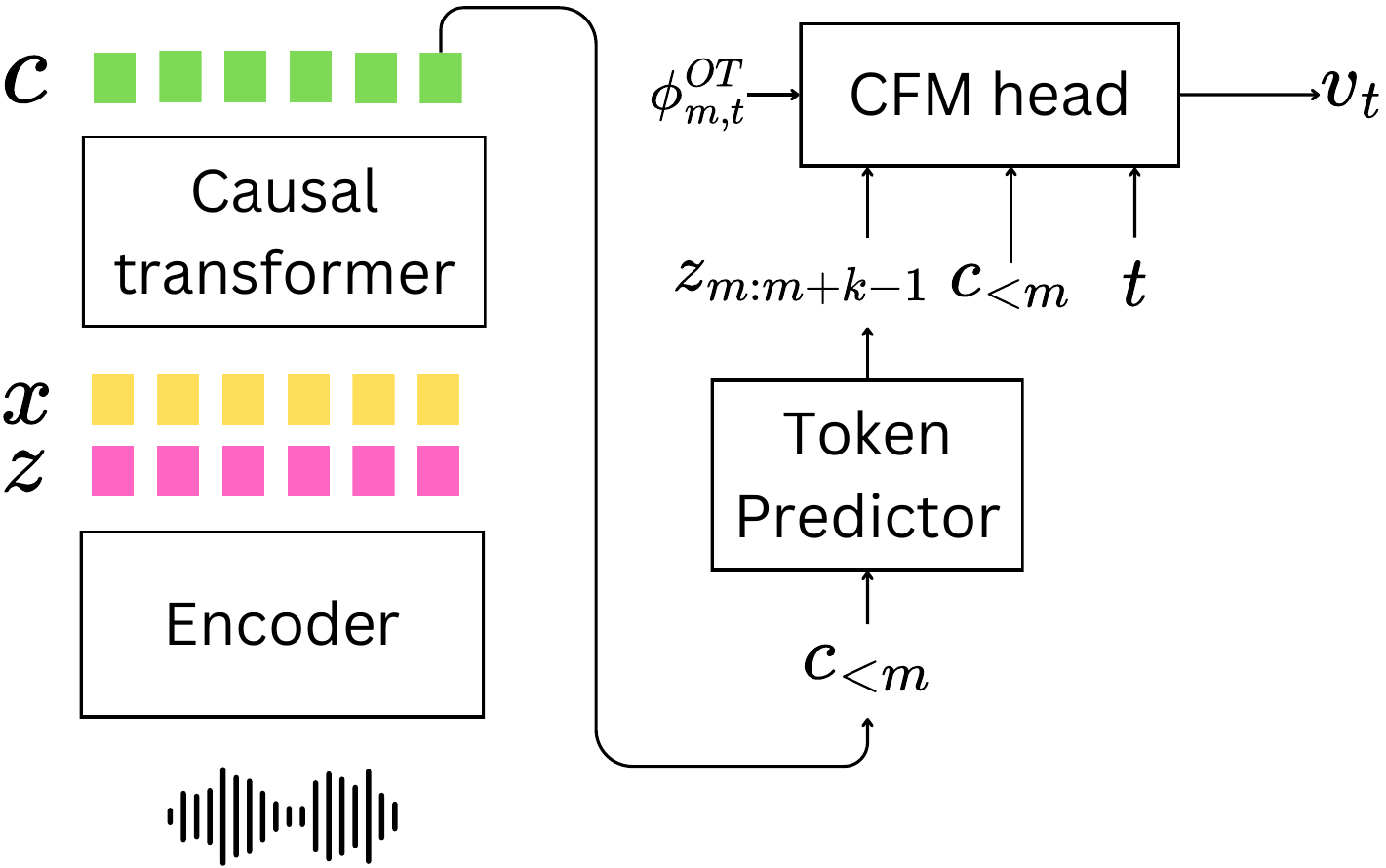}
    \hspace{.07\linewidth}
    \includegraphics[width=0.35\linewidth]{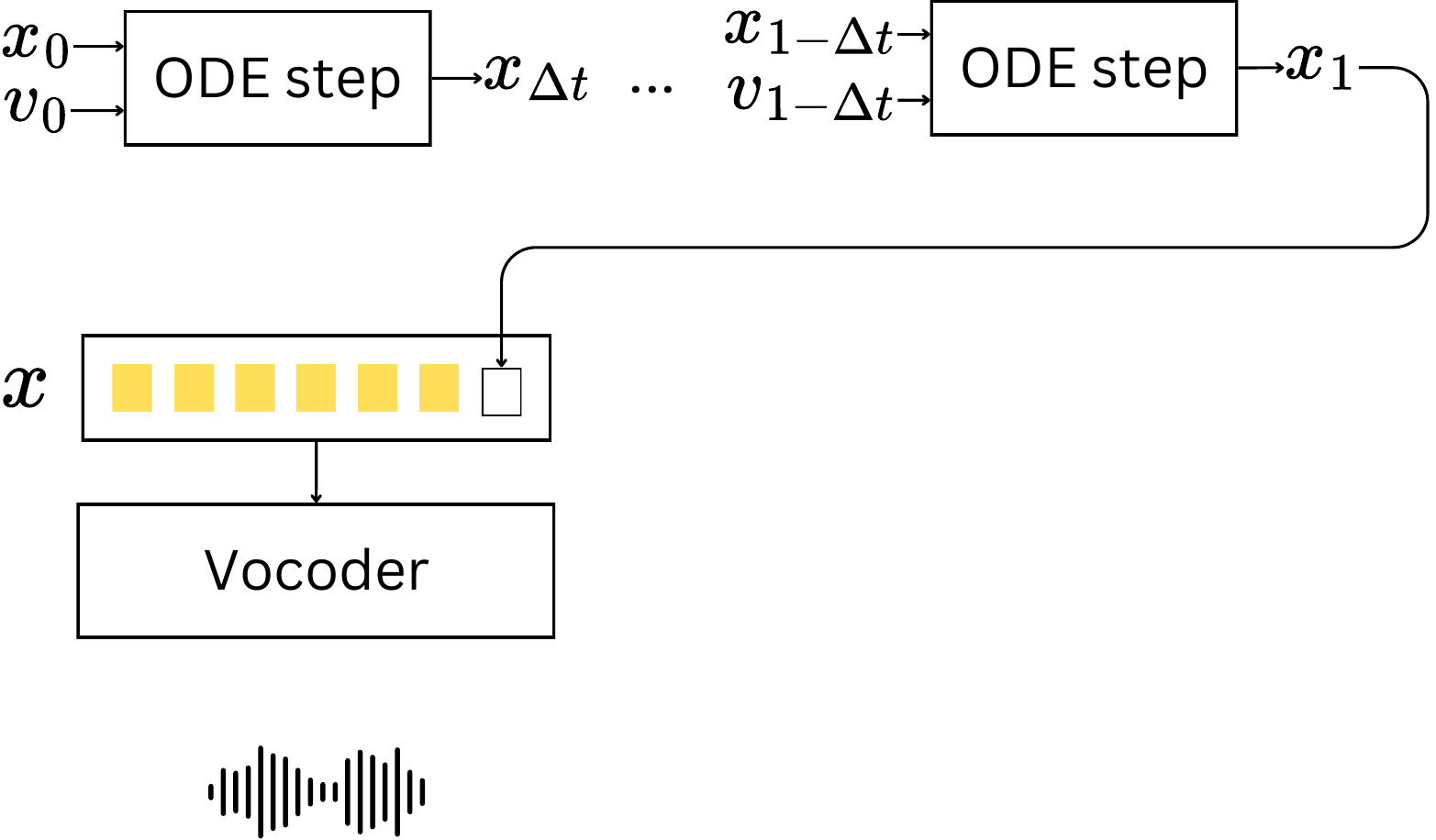}
 \caption{{\it Left}:  The Flow-SLM architecture (see Sec.~\ref{sec:fm}). The encoder maps the waveform into semantic tokens $z$ and continuous embeddings $x$. The causal transformer maps the sequence of embeddings $x$ into a context vector $c$. At each timestep $m$, the token predictor predicts the semantic tokens $z_{m:m+k-1}$ given the context vector $c_{<m}$. The CFM head predicts the vector field $v_t$ from optimal transport conditional flow $\phi^{OT}_{m,t}$ (Sec.~\ref{sec:fm}), conditioning on the semantic tokens $z_{m:m+k-1}$ and the context vector $c_{<m}$.  {\it Right}: The inference process per timestep for the CFM head. During inference, at each timestep, an embedding for the current timestep is generated with an ODE solver. The ODE solver iteratively takes $x_t$ and the velocity from the CFM head ($v_t$) to generate the next $x_{t+\Delta t}$.  After the whole embedding sequence is produced, it is decoded to a waveform by the vocoder.}
    \label{fig:arch}
\end{figure*}
\section{Introduction}

Speech carries both linguistic information---related to the content in the spoken word sequence---and acoustic information, such as the speaker identity, prosody,\footnote{Strictly speaking, prosody can include both linguistic and non-linguistic information.} and noise conditions.  Acoustic information plays an important role both in the naturalness of the signal and by conveying extra-linguistic information such as emotion.

Textless spoken language models (SLMs)~\cite{hassid2023textually,lakhotia2021generative,borsos2023audiolm,kharitonov2021text} aim to directly learn generative models of speech without using text as an intermediate representation. 
Existing approaches primarily focus on generating ``semantic tokens'' that encode linguistic content, while relegating the generation of acoustic information to a separate conditional vocoder. 

In this work, we develop and investigate a new approach to jointly learn from and generate both linguistic and acoustic signals. Joint modeling enables access to and control over both linguistic and acoustic information, allowing the model to capture fine-grained acoustic nuances while maintaining the intended linguistic content.


We propose to directly generate a real-valued vector at each timestep, using a model learned with a conditional flow matching (CFM) objective~\cite{lipman2022flow}. We first generate semantic tokens, and then use a CFM prediction head conditioned on the context and the semantic tokens to predict the current real-valued vector (\autoref{fig:arch}). The generated vectors are then transformed into a waveform by a vocoder.  While the semantic tokens capture linguistic content, the CFM head generates fine-grained acoustic detail. 
In contrast to predicting multiple hierarchical discrete tokens as in residual vector quantization (RVQ) tokenizers~\cite{zeghidour2021soundstream,defossez2024moshi,zhang2024speechtokenizer}, CFM uses a simple regression objective during training, 
substantially reducing 
overhead and memory consumption (see Sec.~\ref{sec:compute}).

We find that simply replacing discrete tokens with continuous-valued tokens 
compromises the generation of meaningful
linguistic content,
because the model learns to focus on local frame-level continuity rather than learning high-level semantic structure.  Instead, we add a prediction head for $N$ future timesteps and find that this approach can mitigate this problem while retaining 
acoustic detail.

Our experimental results show that Flow-SLM achieves comparable performance in terms of semantic metrics to existing discrete-token-only models, while being trained on less compute and data. Flow-SLM is also more acoustically faithful to the ground truth, in terms of distributional metrics and speaker preservation, and is capable of generating diverse acoustic attributes in prompted generation. 
Overall, our approach has
 promising performance simultaneously on multiple metrics of both linguistic and acoustic quality, using relatively modest compute and data resources. 

\section{Related work}
\subsection{Speech tokenization}
Textless SLMs generally require speech tokenization to convert speech signals into discrete tokens. For example, GSLM~\cite{lakhotia2021generative} applies k-means clustering to HuBERT~\cite{hsu2021hubert} representations to extract semantic tokens, which are designed to capture only the linguistic content of the speech rather than enable reconstruction. To address the lack of acoustic detail,~\cite{polyak2021speech} proposes adding speaker and prosody information through a conditional vocoder.

Another approach to tokenization is using a speech codec, 
which encodes speech into discrete tokens and uses a decoder to reconstruct the original waveform, thereby retaining acoustic detail by design.
One commonly used type of codec is based on residual vector quantization (RVQ), used in SoundStream~\cite{zeghidour2021soundstream}, Encodec~\cite{defossez2022high}, and others.  RVQ quantizes the residuals of vector-quantized representations. This allows each frame to be tokenized into $N$ levels of tokens, from coarse to fine. RVQ tokenizers generally yield higher speech quality than conditional vocoders that condition on the semantic tokens to generate waveforms, and the resulting tokens are commonly referred to as acoustic tokens.
To integrate both semantic and acoustic information, SpeechTokenizer~\cite{zhang2024speechtokenizer} and Mimi~\cite{defossez2024moshi} distill semantic tokens into the first level of an RVQ quantizer. This approach produces a unified tokenizer that exploits semantic tokens for modeling linguistic content and acoustic tokens for high-fidelity reconstruction.

\subsection{Spoken language models}
Spoken language models (SLMs)~\cite{arora2025landscape} can be categorized into three main types: (1) speech-aware text LMs~\cite{tang2023salmonn,chu2023qwen}, which can take speech and text as input but produce only text as output, (2) pure speech LMs, where both input and output are speech, which are trained without text~\cite{lakhotia2021generative,borsos2023audiolm,hassid2023textually}, and (3) joint speech-text LMs, which model both speech and text input and output~\cite{chou2023toward,defossez2024moshi,nguyen2025spirit}. Our work falls into the second category.

In pure speech LMs, acoustic details in the generated speech are typically added using a conditional vocoder~\cite{lakhotia2021generative,polyak2021speech,borsos2023audiolm} or via hand-crafted features such as pitch~\cite{kharitonov2021text,nguyen2025spirit} to generate waveforms from a semantic token sequence. However, these approaches offer limited control over acoustic properties based on content and lack access to acoustic information in the input speech during generation. 

Some joint speech-text LMs introduce text as an intermediate representation~\cite{defossez2024moshi,xie2024mini}. These models simultaneously generate text and speech streams. 
Although this approach tends to produce more semantically coherent speech, it requires transcriptions to train the language models. 

\subsection{Conditional flow matching}\label{sec:background_fm}
Conditional flow matching (CFM) is an objective for learning generative models, typically for continuous data, which has been increasingly used in the last few years~\cite{lipman2022flow,liu2022flow}. CFM is one of several types of generative models (another being diffusion models~\cite{song2019generative,ho2020denoising}) that generate by mapping samples from a simple prior distribution to samples from the desired distribution.  CFM learns an ordinary differential equation (ODE), which we can sample from by moving along the learned vector field, parameterized by a neural network, from the prior distribution to the data distribution. CFM has been shown to reduce the number of function evaluations (NFE), the number of times the neural function is evaluated, compared to diffusion models~\cite{lipman2022flow}.
Since CFM learns directly from continuous distributions, it does not require a quantizer to learn a generative model. This makes it well-suited for modalities that are inherently continuous, such as image and audio. 

CFM has been used for speech generation in several ways~\cite{mehta2024matcha,liu2025uniwav,liu2023generative,le2023voicebox,yuan2024continuous}. ~\cite{liu2023generative,liu2025uniwav} use CFM as a pre-training objective for generative speech representation models that can be fine-tuned on other tasks. ~\cite{mehta2024matcha,le2023voicebox} use CFM as an objective for text-to-speech (TTS) synthesis. Our work is, to our knowledge, the first to apply CFM in the context of textless spoken language modeling.

\section{Method}\label{sec:fm}
\subsection{Background: Conditional flow matching (CFM)}\label{sec:background}
In this section, we provide background on CFM~\cite{lipman2022flow} using unconditional generation as an example, but in our models we use a conditional version (\autoref{eq:arc_flow}), i.e. conditioning on the semantic tokens and the past context.

Suppose we have an unknown $d$-dimensional data distribution $q(x)$ that we would like to model. Flow matching (FM) formulates the problem as learning the {\it flow} from a known prior distribution $p_0(x)$ to a distribution $p_1(x)$ that approximates the data distribution $q(x)$: $p_1(x) \approx q(x)$. 
The flow $\phi_t(x)$ is defined by an ODE:
\begin{equation}\label{eq:ode}
\frac{d}{d t} \phi_t(x) = u_t(\phi_t(x)); \phi_0(x)=x,
\end{equation}
where $u_t(\phi_t(x)):  [0,1] \times \mathbb{R}^{d} \to \mathbb{R}^{d}$ is a time-dependent vector field. We can sample $x$ by solving \autoref{eq:ode} from $t=0$ to $t=1$ as shown in the right of \autoref{fig:arch}.

The goal of FM is to learn a parameterized neural function $v_t(x;\theta)$ to match the ground truth vector field $u_t(x)$, minimizing
\begin{equation}
    L_{FM}=\E_{t,p_t(x)}[\Vert  v_t(x;\theta) - u_t(x)\Vert^2].
\end{equation}
However, the ground truth vector field $u_t(x)$ is unknown. Fortunately,~\cite{lipman2022flow} shows that we can 
equivalently learn from a conditional, sample-specific vector field $u_t(x|x_1)$ with $x_1 \sim q(x)$, 
using the CFM loss: 
\begin{equation}
L_{CFM}(\theta)=\E_{t,q(x_1),p_t(x|x_1)}\Vert v_t(x;\theta) - u_t(x|x_1)\Vert^2,
\end{equation}
We use the optimal transport conditional flow~\cite{lipman2022flow}:
\begin{equation}
    \phi_{t}^{OT}(x_0) = t x_1 + (1 - (1 - \sigma_{min})t) x_0,
\end{equation}
with $x_1 \sim q(x_1), x_0 \sim p_0(x)=\mathcal{N}(0, \mathbf{I})$
and $\sigma_{min}=10^{-5}$ following~\cite{liu2023generative}. The corresponding conditional vector field is given by :
\begin{equation}
\label{eq:u_t}
    u_{t}(\phi_{t}^{OT}(x_0)|x_1)=x_1 - (1 - \sigma_{min}) x_0.
\end{equation}
Finally, the learned vector field $v_t(\cdot;\theta)$ is trained to approximate the conditional vector field after re-parameterizing $p_t$:
\begin{equation}
\mathbb{E}_{t,q(x_1),p_0(x_0)}[\Vert v_t(\phi_t^{OT}(x_0);\theta) - \left(x_1 - (1-\sigma_{min})x_0\right) \Vert^2_2].
\end{equation}

In summary, the CFM head takes a sample $\phi_t^{OT}(x_0)$ as input and outputs a prediction of the vector field $v_t(\cdot;\theta)$ to approximate the conditional vector field $u_{t}(\phi_{t}^{OT}(x_0)|x_1)$ during training. 

\subsection{Flow-SLM architecture and training}

Our architecture consists of four components: an encoder, a causal transformer, a semantic token predictor, and a CFM head as shown in~\autoref{fig:arch}. For the encoder we use a pre-trained, frozen encoder, 
resulting in a set of learnable parameters $\theta=\{\theta_{T},\theta_{sem},\theta_{CFM}\}$ for the causal transformer, semantic token predictor, and CFM head respectively. 

Given a distribution over speech waveforms $\mathcal{D}$,  we sample a waveform of length $L$ samples: $s_{1:L} \sim \mathcal{D}$. The encoder $\Enc\cdot$ transforms the waveform into a sequence of $M$ embeddings $x_{1:M} \in \mathbb{R}^{M \times d}$, and a semantic token sequence $z_{1:M}$  where each $z_m \in \mathcal{V}$ for some vocabulary $\mathcal{V}$:
\begin{equation}
    x_{1:M}, z_{1:M} =\Enc{s_{1:L}}.
\end{equation}
We use the pre-quantization embedding in Mimi~\cite{defossez2024moshi}, i.e. the output of the encoder, as our embedding $x$, and its first RVQ layer as semantic tokens $z$. 

The causal transformer $\T{\cdot}$ encodes the past context $x_{1:m-1}$ into a context vector $c_{<m}$: 
\begin{equation}
c_{<m} = \T{x_{1:m-1}}.
\end{equation}
The semantic token predictor $P_{sem-k}(\cdot)$ 
predicts the next $k$ semantic tokens $z_{m:m+k-1}$ given the context vector $c_{<m}$, and is learned by minimizing a {\it multi-token} cross-entropy loss:
\begin{equation}
   L_{sem-k}(\theta)=\frac{1}{k}\E[-\log P_{sem}(z_{m:m+k-1}|c_{<m}) ]
\end{equation}
As we demonstrate in our experiments, predicting further into the future is crucial for achieving good performance in terms of content-related metrics with our model.

The CFM head is responsible for modeling the distribution of $x_m$ at each timestep $m$, given the previous context and the predicted semantic tokens.  
Our flow matching head $v_t(\cdot;\theta_{CFM})$ is trained with a CFM objective as described in Sec. \ref{sec:background}, 
but with additional conditioning on the context vector $c_{<m}$ and the semantic tokens $z_{m:m+k-1}$:
\begin{eqnarray}\label{eq:arc_flow}
    &L_{CFM-k}(\theta)&\\
    &=\mathbb{E}_{t,q(x_1),p_0(x_0)}&[\Vert v_t(\phi_t^{OT}(x_0), c_{<m}, z_{m:m+k-1}) - u_t \Vert^2_2], \nonumber
\end{eqnarray}
where $u_t$ is defined as in~\autoref{eq:u_t} and where we have dropped the subscript $m$ in $x_m$ for simplicity. 
Our final loss is
\begin{equation}
    L(\theta)=L_{sem-k}(\theta) + L_{CFM-k}(\theta)
\end{equation}

\subsection{Flow-SLM inference}
During inference, we first generate the context vector $c_{<m}$ based on the available context and sample the semantic token  $z_{m:m+k-1}$. Then at each timestep $m$, we sample from the prior distribution $x_{m,0} \sim \mathcal{N}(0, \mathbf{I})$, then use the learned ODE produced by the CFM head, which is conditioned on $\{c_{<m}, z_{m:m+k-1}\}$, to transform $x_{m,0}$ into $x_{m,1} = x_m$. 
The generated embedding $x_{m}$ is appended to the input of the causal transformer for predicting
the following timestep $m+1$. After collecting the whole sequence until the end-of-sequence token, we decode the embedding sequence into a waveform using a vocoder (the Mimi decoder + quantizer) as shown in~\autoref{fig:arch}.

We note that Flow-SLMs require only one forward pass for the causal transformer, but multiple passes for the CFM head, 
to generate one token, saving the compute for the transformer. 


\subsection{Comparison between RVQ and CFM}\label{sec:compute}
Another approach for modeling acoustic details is to use RVQ tokens~\cite{defossez2024moshi}. RVQ has the disadvantage that it requires multiple levels of acoustic tokens to be ordered in some way, and this can lead to a larger computation overhead when predicting them auto-regressively. 
For example, Mimi~\cite{defossez2024moshi} represents speech as 12.5 real-valued vectors per second,
each of which is quantized into 32 tokens ordered from coarse to fine. 
To model the hierarchical structure of these tokens during generation, they are arranged either by using a ``delay pattern''~\cite{defossez2024moshi} or by flattening them into a sequence~\cite{borsos2023audiolm}.

Training with the CFM objective requires less memory than its RVQ counterpart. For example, in RVQ with token flattening, training with $N$ levels of tokens requires memory of $B * T * N * \vert \mathcal{V}\vert$ per batch, where $B$ is the batch size, $T$ is the number of timesteps, and $V$ is the vocabulary.  
On the other hand, flow matching requires running only one $d$-dimensional linear regression head per timestep ($B * T * d$).


\section{Experiments}
\subsection{Setup}
Our main training data is the English subset of Multilingual LibriSpeech (MLS-En)~\cite{pratap2020mls},  
which contains approximately 45k hours of speech, corresponding to around 2 billion continuous tokens with the Mimi encoder\cite{defossez2024moshi}. We also use an extended training set, which includes these 45k hours and an additional 20k hours from the clean subset of The People’s Speech~\cite{galvez2021people} to expose the model to a non-audiobook domain. We use MLS-En (45k) as the base setup if not otherwise specified, and MLS-En plus The People’s Speech clean (65k total) as the extended setup, denoted by "-ext". 

We initialize the causal transformer from a text LM, following TWIST~\cite{hassid2023textually}, specifically using two OpenELM~\cite{mehta2024openelm} checkpoints (OpenELM-270M and OpenELM-1B) for our two model sizes. For the CFM head, we adopt the same architecture as in~\cite{li2024autoregressive} 6 residual blocks ($\sim$150M parameters). 

The model is trained for one epoch with a batch size of 128 utterances. We use 8-bit AdamW~\cite{dettmers20218} as the optimizer to reduce GPU memory usage. We use the Mimi encoder~\cite{defossez2024moshi} to extract representations, with pre-quantizer outputs serving as the embeddings $x$ and the first RVQ level used as the semantic tokens $z$. We use classifier-free guidance (CFG)~\cite{ho2022classifier} to sharpen the distribution,  by randomly dropping the conditioning input to the CFM head $(z_{m:m+k-1}, c_{<m})$ with a probability of 0.05 during training.

During inference, we use nucleus sampling~\cite{holtzman2019curious} with a top-$p$ value of 0.95 when sampling semantic tokens. We find that Flow-SLMs tend to generate excessive silence, which we reduce by
penalizing the logits of silence tokens during sampling.\footnote{Silence tokens are identified using VAD from the TorchAudio package~\cite{hwang2023torchaudio}, and we subtract 10 from their logits for decoding steps.}
We sample from the learned ODE using a Euler ODE solver with NFE=32 based on preliminary experiments. We apply a temperature of 0.8 to the prior Gaussian distribution (which is equivalent to changing the variance of the prior) and set the CFG scale~\cite{ho2022classifier} to 0.3.

The generated embeddings are converted to waveforms using the Mimi quantizer with the first 16 RVQ levels (including the later levels introduces more artifacts) and decoder~\cite{defossez2024moshi}.

\subsection{Evaluation}

\paragraph{Semantic modeling}  We evaluate semantic modeling performance using sWUGGY and sBLIMP from the ZeroSpeech 2021 challenge~\cite{nguyen2020zero}. sWUGGY measures the model’s ability to distinguish words from non-words: each testing pair consists of a word and a non-word, and the pair is considered correct if the model assigns a higher probability to the true word than to the non-word. We use all the vocabulary in the sWUGGY experiment. sBLIMP follows a similar setup but focuses on sentence grammar.

\paragraph{SALMon metrics}  We measure acoustic consistency, semantic-acoustic alignment, and background alignment using the SALMon~\cite{maimon2025salmon} benchmark. One important note is that we use the CFM head to generate acoustic details conditioned on the semantic tokens, and the CFM head does not provide an explicit likelihood to evaluate the acoustic aspect of the generated data. Therefore, we only use the semantic token predictor to measure likelihood in SALMon, which may not fully reflect the model's ability on acoustic tasks. 

For evaluation of semantic-acoustic alignment, we use SALMon's sentiment alignment task. In this task, a ``positive" utterance conveys the same sentiment in both the content and the prosody (for example, using a sad tone when saying ``I am sad"), while a negative example is the same sentence with mismatched sentiment. For acoustic consistency, we average over all attributes in SALMon. Acoustic consistency includes pairs of positive and negative samples with the same content. Positive samples have consistent attributes, such as speaker, while negative samples are inconsistent at word boundaries. 

\paragraph{Generated speech quality}  To evaluate the quality of the generated speech, we perform prompt continuation using a randomly sampled subset of 500 utterances 
from the LibriSpeech test-clean and test-other sets~\cite{panayotov2015librispeech}. We use the first 3 seconds of each utterance as a prompt and continue generation until 10 seconds. We sample 4 continuations for each prompt and use these continuations for all generation-related metrics. We then evaluate in several ways.  First, we transcribe the speech using Whisper-large-v3-turbo~\cite{radford2023robust} and evaluate the perplexity of the text using LLaMA-3.2-1B~\cite{grattafiori2024llama}, a metric referred to as \textit{genPPL}~\cite{maimon2025slamming,hassid2023textually}.\footnote{Note that the text perplexity here measures only the fitness of the generated speech content to LlaMA-3.2-1B's distribution, which may not fully align with human 
preferences. We use the small LM for genPPL following prior work~\cite{maimon2025slamming,hassid2023textually}, but a better language model may provide better evaluation.}

In addition, we measure the speaker similarity between the prompt and continuation, 
using a WavLM-large-TDNN~\cite{chen2022wavlm} speaker verification model, which is fine-tuned from WavLM, 
following~\cite{le2023voicebox,he2024emilia,liu2025uniwav}.\footnote{\url{https://github.com/microsoft/UniSpeech/tree/main/downstreams/speaker_verification}}. For this purpose, we resynthesize the prompt with the Mimi decoder~\cite{defossez2024moshi} to eliminate any inconsistency due purely to the vocoder.

Finally, we use the Fréchet speech distance (FSD)~\cite{le2023voicebox,he2024emilia} to measure the dissimilarity between the generated distribution and the ground-truth distribution for prompted generation. Specifically, we use two variants of FSD:  e2v FSD uses the utterance-level representations from emotion2vec~\cite{ma2023emotion2vec} to measure the expressiveness of the generated speech, following~\cite{he2024emilia}, and wlm FSD uses WavLM-large layer 6 (the layer is selected based on a previous study~\cite{baas2023voice}) as features to measure the overall quality of the per-frame distribution. 

\section{Results}
For all of the results below, unless otherwise specified, we train Flow-SLMs by predicting $4$ future tokens (i.e., using the $L_{sem-4}$ loss) and using them as a condition for the CFM head.
\subsection{Comparison to other SLMs}
\begin{table}[t]
\centering
\caption{ ZeroSpeech metrics. The table is divided into sections corresponding to models of approximately the same size. 
 *: AudioLM only uses the semantic token model (300M parameters) for sWuGGY and sBLIMP. }\label{tab:likelihood}
\begin{tabular}{ll@{\hspace{1em}}cc}
\hline
\# & Model &sWUGGY$\uparrow$ & sBLIMP$\uparrow$  \\
\midrule
1& TWIST-350M &  69.7 & 56.2 \\
2& AudioLM-300M* & 71.5 & 64.7 \\
3& Flow-SLM-270M  &  68.7 & 57.3    \\

\midrule
4&TWIST-1.3B  &  72.7 & 57.0  \\
5&VoxtLM-1.3B  &65.6 & 57.1 \\
6&Flow-SLM-1B &  69.8 & 60.0  \\
7&Flow-SLM-1B-extend &  65.7 & 60.9  \\
\midrule
8&Moshi-7B & 74.3 & 58.9   \\
9&SpiritLM-base-7B& 69.0 & 58.3 \\
10&SpiritLM-exp-7B & 65.0 & 54.2 \\

\end{tabular}
\end{table}
\begin{table}[t]
\caption{Metrics related to generation. 
 ``wlm FSD'' = wavlm FSD, ``spk sim'' = speaker similarity, ``ext.'' = extended. The ablation study (the last three rows) are conducted on the dev set. }\label{tab:gen}
\centering
\begin{tabular}{l@{\hspace{1em}}rrrr}
\hline
Model &  genPPL$\downarrow$ & spk sim$\uparrow$ & e2v FSD$\downarrow$ & wlm FSD$\downarrow$  \\
\midrule
ground truth &64.3& 0.66  \\
\midrule
TWIST-350M & 116.7 & 0.09 & 6.16 & 1416.7 \\
Flow-SLM-270M & 173.3 & 0.36 & 3.23 & 1235.4  \\
\midrule
TWIST-1.3B & 98.6 & 0.09 & 5.62 & 1417.4 \\
Flow-SLM-1B  & 147.5& 0.43 & 2.82 & 1018.1\\
Flow-SLM-1B ext. & 137.6 & \textbf{0.45} & \textbf{2.62} & \textbf{997.9} \\
\midrule
SpiritLM-base-7B & \textbf{54.7} & 0.11 &5.31 &1954.8  \\
SpiritLM-exp-7B & 84.0 & 0.11 &3.85 & 1315.1  \\
\midrule\midrule
Flow-SLM-270M & 178.9 & 0.37 & 4.70 & 1240.9  \\
\hspace{1em}- future prediction & 314.1 & 0.37 & 6.20 & 1007.1 \\
\hspace{2em}- sem. token & 407.1  &  0.42 &  5.56 & 1045.6 \\

\end{tabular}
\end{table}

We compare Flow-SLM to previous approaches, including TWIST~\cite{hassid2023textually}, SpiritLM~\cite{nguyen2025spirit}, Moshi~\cite{defossez2024moshi}, AudioLM~\cite{borsos2023audiolm}, and VoxtLM~\cite{maiti2024voxtlm} in~\autoref{tab:likelihood} and~\autoref{tab:gen}.  First, we consider linguistic content-related metrics, including sWUGGY and sBLIMP in~\autoref{tab:likelihood} and genPPL in~\autoref{tab:gen}. We can see that Flow-SLM-1B lags behind TWIST-1.3B~\cite{hassid2023textually} in terms of sWUGGY, but achieves a better result on sBLIMP than all previous models except for AudioLM. Flow-SLM-1B outperforms the larger SpiritLM-base-7B in terms of both sWUGGY and sBLIMP. When scaling up the data size, Flow-SLM-1B-extend has a slightly improved sBLIMP but a degraded sWUGGY. One possible explanation is that adding The People's Speech as training data introduces a distribution mismatch between the training data and the testing data in sWUGGY, as half of the vocabulary in sWUGGY is taken from the LibriSpeech training data.

\begin{figure}
    \centering
    \includegraphics[width=\linewidth]{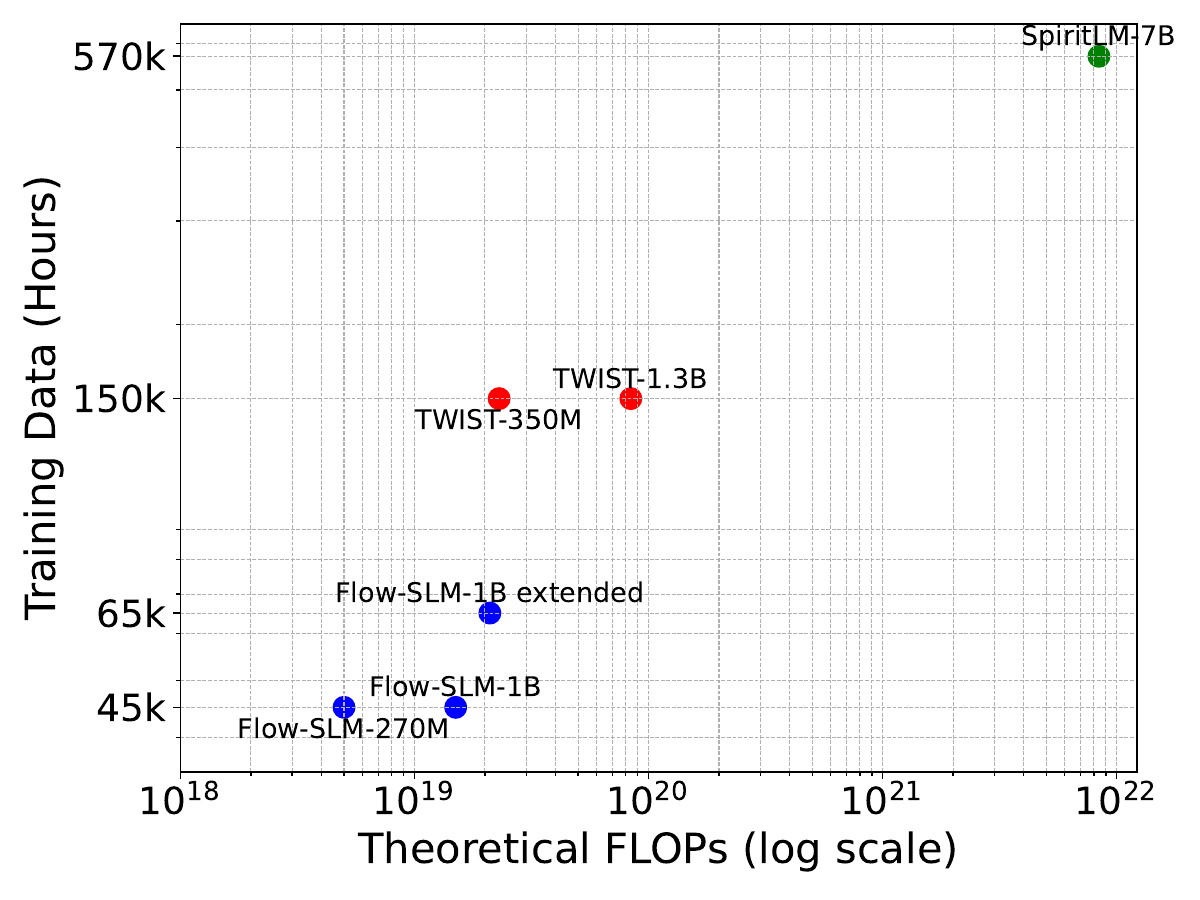}
    \vspace{-.3in}
    \caption{Compute and data comparison across models. Following~\cite{hoffmann2022training}, we estimate the theoretical FLOPs as $6
*N_{param}*D_{tokens}$, where $N_{param}$ is the number of training parameters excluding embeddings and $D_{tokens}$ is the number of training tokens. For SpiritLM, we include the total amount of speech data, including both speech-only and speech-text data.}
    \label{fig:compute}
\end{figure}

However, in terms of genPPL, the Flow-SLM-1B family performs worse than the TWIST and SpiritLM~\cite{nguyen2025spirit} families, suggesting that Flow-SLMs are less coherent in terms of the generated linguistic content. 
Potential reasons include: (1) Flow-SLMs are trained with less compute and data (see~\autoref{fig:compute}), and (2) the joint learning of linguistic and acoustic information may compromise the modeling of longer-range semantics. An important direction for future work is therefore to analyze the scaling behavior of Flow-SLMs.
\begin{table}[t]
\caption{Salmon benchmark accuracy (\%). The acoustic consistency metric is an average of all acoustic consistency categories in Salmon. "sent.~alignment" stands for sentiment alignment, and "bg.~alignment" stands for background alignment. *: We count the parameters based on the checkpoint provided.}\label{tab:salmon}
\centering
\begin{tabular}{lc@{\hspace{1em}}ccc}
\hline
 Model & Consist. $\uparrow$ & sent. alignment $\uparrow$ & bg. alignment $\uparrow$  \\
\midrule
TWIST-350M &  62.5 & 51.5 & 56.5 \\
Flow-SLM-270M  &  70.8 & \textbf{60.0} &	55.5 \\

\midrule
TWIST-1.3B  &  62.5 & 53.0 & 56.5 \\
Flow-SLM-1B &  \textbf{71.6} & 57.0&56.0 \\
Flow-SLM-1B-ext. &  \textbf{71.6} & 57.0 &	53.0 \\
\midrule
pGSLM-151M* & 64.8 & 55.5 &53.5\\
TWIST-7B & 63.3 & 51.5 & 54.5 \\
SpiritLM-exp-7B & 69.0 & 52.0 & \textbf{59.5} \\

\end{tabular}
\end{table}


Next, we use the SALMon benchmark~\cite{maimon2025salmon} to measure acoustic consistency, sentiment alignment, and background alignment, as shown in~\autoref{tab:salmon}.  We compare only to the top performing models on the SALMon benchmark~\cite{maimon2025salmon}. Flow-SLMs outperform TWIST and SpiritLM models in terms of acoustic consistency and sentiment alignment, but perform slightly worse in terms of background alignment. 
The poorer results on background alignment may be caused by the fact that we train our models mainly on MLS, which contains only audiobooks, and therefore is not exposed to a wide variety of acoustic backgrounds, whereas TWIST and SpiritLM are.

In terms of speaker similarity, we can see from~\autoref{tab:gen} that, when TWIST and SpiritLM do not have access to a conditional vocoder (conditioning on the prompt speaker), they fail to maintain speaker consistency for the continuation. Flow-SLMs, on the other hand, are better able to retain speaker similarity. In terms of e2v FSD and wlm FSD (\autoref{tab:gen}), Flow-SLMs again outperform SpiritLM and TWIST.

\subsection{Does detailed acoustic context hurt semantic learning?}
\begin{table}[t]
\caption{Ablation study on the dev set of ZeroSpeech and validation loss on the LibriSpeech dev set. ``Token input'' means the model takes discrete input, and ``vector'' input means the model takes continuous vector input. $L_{sem-k}$ refers to the loss for predicting $k$ future tokens. $L_{CFM-N}$ indicates that the CFM head conditions on $N$ predicted future tokens.}\label{tab:zerospeech}
\centering
\begin{tabular}
{cc@{\hspace{1em}}c@{\hspace{-.3em}}ccc}
\hline
\# & input  & Objective & sWUGGY$\uparrow$ & sBLIMP$\uparrow$ & CE loss \\
\midrule
\multicolumn{3}{c}{TWIST-350M} & 70.2 & 55.2 \\
\midrule
1 & token & $L_{sem-1}$ &  69.3 &  55.7 & 3.44\\
2 & token & $L_{sem-2}$ &  70.0 & 57.1 & 4.16 \\
3 & token & $L_{sem-4}$ &  63.7 & 53.3 & 4.94 \\
4 & vector & $L_{sem-1}$ &  63.7 & 52.4 & 2.28 \\
5 & vector & $L_{sem-2}$ &  68.1 & 55.5 & 3.17 \\
6 & vector & $L_{sem-4}$ &  68.6 & 57.8 & 4.27  \\
7 & vector & $L_{sem-1}$ +  $L_{CFM-1}$ &  67.5 & 57.0 & 4.27\\
8 & vector & $L_{sem-4}$ + $L_{CFM-4}$ &  68.3 & 57.1 &  4.23\\

\end{tabular}
\end{table}
We conduct an ablation study on the dev set of the ZeroSpeech benchmark~\cite{nguyen2020zero} and the LibriSpeech dev set to investigate whether using fine-grained acoustic context (using vector $x$ instead of token $z$) compromises semantic understanding. As shown in row 1 of \autoref{tab:zerospeech}, using standard CE loss ($L_{sem-1}$) and Mimi semantic tokens as input produces comparable performance on sWUGGY and sBLIMP to TWIST-350M~\cite{hassid2023textually}. However, when we replace the input with continuous vectors (row 4), performance drops despite achieving a lower CE validation loss than row 1. This indicates that the model becomes better at predicting the next semantic token but fails to capture long-term dependencies, which are crucial for learning the lexicon, higher-level semantics, and syntax.
We attribute this finding to the continuity of speech signals across time:  When the context includes both semantic and acoustic information, the model can exploit acoustic continuity to predict the next semantic token easily. 

To address this issue, we have proposed predicting additional future tokens. For example, using the $L_{sem-4}$ objective to predict four future tokens (row 6) improves performance on both sWUGGY and sBLIMP. Predicting more future tokens helps in both token and vector input settings, but provides more benefit with vector inputs, as it mitigates their tendency to account only for local continuity. On the other hand, even predicting two future tokens helps the token-only model better capture lexical and grammatical patterns.  This finding is an interesting by-product of our proposed future prediction loss, which may be interesting to study even in token-only models.

From the genPPL results in \autoref{tab:gen}, we observe that predicting further into the future and conditioning on future tokens improves generation perplexity (genPPL) significantly. Conversely, models that do not use semantic tokens tend to focus solely on local information, leading to poor genPPL.

\subsection{Does the CFM prediction head hurt semantic learning?}

When adding a CFM head to the model (row 7 in~\autoref{tab:zerospeech}), we observe a drop in semantic metrics compared to row 6. This suggests that the model must encode additional acoustic information into the context vector to learn to generate acoustic details, which in turn reduces its capacity to retain semantic information. However, providing more conditioning (row 8) helps to mitigate this issue. Therefore, we adopt $L_{sem-4} + L_{CFM-4}$ as the default objective for Flow-SLM.

This performance drop is consistent with the findings of~\cite{nguyen2025spirit}, where SpiritLM-expressive, which includes prosody and style tokens, performs worse than SpiritLM-base on semantic metrics (as shown also in~\autoref{tab:likelihood} and~\autoref{tab:gen}).

\subsection{Is acoustic information degraded when predicting future semantic tokens?}

The analysis above shows that predicting semantic tokens further into the future helps to retain semantic information.
The next question we ask is whether we lose acoustic information when emphasizing semantic information in this way.

From the speaker preservation results in \autoref{tab:gen}, we see that when we learn solely from the CFM head (Flow-SLM-270M - future prediction - sem. token), the model has better speaker similarity compared to the models that use semantic tokens. 
For e2v FSD, which is used to evaluate expressiveness, we find that Flow-SLM works better with future prediction. We suspect that this is because expressiveness is tied to the semantic content.
On the other hand, the results for wlm FSD, which focuses on the frame distribution similarity alone, show that the generated samples are closer to the ground truth distribution without future prediction.

\section{Conclusion and future work}
Our proposed Flow-SLM models jointly learn linguistic and acoustic information for textless spoken language models.  Unlike prior approaches that require separate conditional vocoders, Flow-SLMs can natively generate acoustically consistent and expressive speech, largely without compromising semantic content.  Our results show that the proposed approach of predicting multiple future tokens helps the model better capture linguistic content.  Overall, our models perform competitively with existing models by many metrics, including both acoustic and linguistic evaluations, while using much less training data and compute, although there is a tradeoff between acoustic detail and semantic content.  Future work on scaling the models up is needed to establish whether scale can close the remaining gaps in terms of semantic modeling.

Our work thus far focuses on the textless setting, which generally performs worse at learning semantics compared to models that generate text as intermediate representations~\cite{defossez2024moshi}. Extending our model to a joint speech and text setup could therefore lead to more coherent generated speech. In terms of acoustic quality, it will be important to compare Flow-SLMs with models that generate acoustic RVQ tokens to better understand the trade-offs\footnote{To the best of our knowledge, there is no publicly available textless SLM based on RVQ tokens.} and to extend the training data to a broader range of acoustic conditions.



\bibliographystyle{IEEEtran}  
\bibliography{ref_new}  
\end{document}